\documentclass[lettersize,journal]{IEEEtran}
\usepackage{amsmath,amsfonts}
\usepackage{algorithmic}
\usepackage{algorithm}
\usepackage{array}
\usepackage[caption=false,font=normalsize,labelfont=sf,textfont=sf]{subfig}
\usepackage{textcomp}
\usepackage{stfloats}
\usepackage{url}
\usepackage{verbatim}
\usepackage{graphicx}
\usepackage{cite}
\usepackage{booktabs}

\usepackage[unicode=true,
 bookmarks=true,bookmarksnumbered=false,bookmarksopen=false,
 breaklinks=false,pdfborder={0 0 1},backref=false,colorlinks=true]
 {hyperref}
\hypersetup{
 linkcolor=magenta, urlcolor=blue, citecolor=blue, pdfstartview={FitH}, unicode=true}

\begin{document}

\title{A Generative Neural Annealer for Black‑Box Combinatorial Optimization}

\author{Yuan-Hang Zhang and Massimiliano Di Ventra,~\IEEEmembership{Fellow, IEEE}
\thanks{Yuan-Hang Zhang and Massimiliano Di Ventra are with the Department of Physics, University of California, San Diego, La Jolla, CA 92093, USA (emails: yuz092@ucsd.edu; diventra@physics.ucsd.edu)}
}



\maketitle

\begin{abstract}
    We propose a generative, end-to-end solver for black-box combinatorial optimization that emphasizes both sample efficiency and solution quality on non-deterministic polynomial time problems. Drawing inspiration from annealing-based algorithms, we treat the black-box objective as an energy function and train a neural network to model the associated Boltzmann distribution. By conditioning on temperature, the network captures a continuum of distributions—from near-uniform at high temperatures to sharply peaked around global optima at low temperatures—thereby learning the structure of the energy landscape and facilitating global optimization. When queries are expensive, the temperature-dependent distributions naturally enable data augmentation and improve sample efficiency. When queries are cheap but the problem remains hard, the model learns implicit variable interactions, effectively ``opening'' the black box. We validate our approach on challenging combinatorial tasks under both limited and unlimited query budgets, showing competitive performance against state-of-the-art black-box optimizers.
\end{abstract}


\section{Introduction}

Many real-world decision problems—from scheduling and routing to circuit design—can be framed as combinatorial optimization: the task of searching over a discrete space that grows exponentially with the number of variables. When the objective function is a black box—accessible only through queries, each of which may be costly (e.g., due to long-running simulations or physical experiments)—the problem becomes one of black-box combinatorial optimization \cite{juels1996topics, alarie2021two}. This setting presents two major challenges: first, the limited budget of function evaluations demands sample-efficient algorithms; second, the combinatorial landscape is often rugged, making it difficult to extract meaningful patterns from the observed data.

According to the ``no free lunch'' theorem \cite{wolpert1997no}, no optimizer is universally superior: averaged over all possible objective functions, any two algorithms perform equally well, provided they have no knowledge of the properties of the objective functions they are trying to extremize. Thus, success hinges on exploiting structure. Real-world optimization problems often exhibit properties like smoothness, modularity, decomposability, and symmetry. Many black-box optimizers implicitly rely on such assumptions to outperform random or uninformed search \cite{alarie2021two, phan2003model}.

When structural information is available, tailored solvers—such as semidefinite programming for convex problems or graph neural networks for graph-structured inputs—can achieve strong performance. In the black-box setting, classical metaheuristics like hill climbing \cite{skiena2008algorithm}, simulated annealing \cite{kirkpatrick1983optimization}, and tabu search \cite{glover1986future} rely on the assumption that the objective function behaves smoothly under local changes. While simple and effective in some cases, these methods often require many queries and are prone to getting trapped in local minima.

Surrogate-based Bayesian optimization (BO) offers an alternative by modeling the objective with a probabilistic surrogate—such as a Gaussian Process or tree-based estimator—and selecting new queries via an acquisition function \cite{mockus2005bayesian, garnett2023bayesian}. BO excels in low-dimensional continuous domains, but its effectiveness diminishes in high-dimensional or discrete spaces. In such settings, surrogate models may over-smooth the landscape, missing sharp transitions common in combinatorial problems. Additionally, BO typically scales poorly with the number of evaluations and requires solving a secondary nonconvex optimization problem to select each new query \cite{baptista2018bayesian, oh2019combinatorial}, which adds significant computational overhead.

Given these considerations, deep learning emerges as a strong candidate for black-box combinatorial optimization. Neural networks offer a powerful and flexible means of approximating high-dimensional search spaces, can capture global patterns from data, and, critically, their training cost does not scale with the number of observations. A wide range of deep learning-based optimizers have been proposed, often tailored to specific families of problems with domain-specific structure \cite{bello2016neural, khalil2017learning, bengio2021flow, sanokowski2024diffusion, zhang2023let, bengio2021machine, karimi2022machine, mazyavkina2021reinforcement, vesselinova2020learning}.

In this paper, we introduce the \textit{Generative Neural Annealer} (GNA), an end-to-end solver for black-box combinatorial optimization built on a decoder-only transformer architecture. When query budgets are limited, GNA leverages the available data to model the structure of the objective function, achieving high sample efficiency without requiring any domain-specific knowledge. Inspired by simulated annealing \cite{kirkpatrick1983optimization}, we treat the black-box objective $f(x)$ as an energy function and train the model to approximate the Boltzmann distribution
\begin{equation}
    p(x,\beta)\propto \exp\big(-\beta f(x)\big), \label{eq:Boltzmann}
\end{equation}
across a continuum of inverse temperatures $\beta \in [\beta_{\text{low}}, \beta_{\text{high}}]$.

Classical algorithms typically simulate Eq.~\eqref{eq:Boltzmann} using Markov Chain Monte Carlo (MCMC) methods. However, mixing times can become exponentially long when the energy landscape is highly rugged or frustrated \cite{geyer1992practical}. Parallel tempering \cite{swendsen1986replica, earl2005parallel} addresses this by simulating multiple replicas of the system at different temperatures, while various nonlocal cluster updates have also been developed to accelerate mixing \cite{houdayer2001cluster, zhu2015efficient}. More recently, machine learning has been leveraged to approximate Eq.~\eqref{eq:Boltzmann} directly: the Boltzmann Generator \cite{noe2019boltzmann} accelerates MCMC by learning a neural approximation of the target distribution, and Variational Neural Annealing (VNA) \cite{hibat2021variational} further extends this idea to perform annealing using the learned distribution.

Our approach is conceptually similar to VNA, but with a key distinction analogous to the difference between parallel tempering and simulated annealing: instead of learning a single distribution and annealing its temperature, we model an entire continuum of distributions across temperatures, learning a smooth interpolation between a high-temperature, near-uniform distribution and a low-temperature, sharply peaked distribution concentrated near global optima.

Training is simple and fully black-box: the model proposes a batch of candidate solutions, evaluates them using the black-box function $f$, and updates its parameters via standard stochastic gradient methods to increase the likelihood of low-energy samples. Because the same network models the full range of temperatures, it inherently captures multi-scale structures and offers natural escape routes from local minima, without relying on external restart heuristics.
 
Our main contributions are summarized as follows:
\begin{itemize}
    \item We propose GNA, a black-box solver for combinatorial optimization that learns a temperature-parameterized Boltzmann distribution to guide the search for optimal solutions.
    \item We design two training regimes for GNA—one tailored to limited-query settings, and one for cases with abundant queries—and demonstrate strong performance in both.
    \item We empirically show that GNA effectively learns the structure of the objective by capturing interactions between input variables, without access to domain knowledge.
\end{itemize}

\section{Related Work}
\subsection{Bayesian optimization (BO)}

Bayesian optimization is a widely used framework for black-box optimization \cite{mockus2005bayesian, garnett2023bayesian}, and has seen success particularly in applications such as hyperparameter tuning \cite{bergstra2013making, wu2019hyperparameter, turner2021bayesian}. Most BO algorithms are designed for continuous input spaces and rely on Gaussian Processes (GPs) as surrogate models. For discrete or combinatorial domains, alternative surrogates are used. For example, BOCS \cite{baptista2018bayesian} fits a sparse quadratic polynomial to model the objective and uses Thompson sampling combined with semidefinite or local search to propose new candidates. COMBO \cite{oh2019combinatorial} constructs a graph-based kernel via the Cartesian product of subgraphs defined over variable domains, enabling GPs to operate in discrete spaces.

While these methods are sample-efficient, they often incur high computational overhead—both in fitting the surrogate model and in optimizing the acquisition function, which itself may be a hard combinatorial problem. To mitigate this, COMEX \cite{dadkhahi2020combinatorial} proposes a multilinear polynomial surrogate to simplify acquisition optimization, and MerCBO \cite{deshwal2021mercer} leverages Mercer features with Thompson sampling to balance tractability and performance. An alternative approach is the Tree-structured Parzen Estimator (TPE) \cite{bergstra2013making, watanabe2023tree}, which models the density of promising vs. non-promising regions in the input space and selects candidates by maximizing expected improvement under these densities. TPE is especially well-suited for categorical or mixed-variable spaces.

\subsection{Learning-Based Combinatorial Optimizers}
Machine learning has become an increasingly popular approach to solving combinatorial optimization problems \cite{bello2016neural, khalil2017learning, bengio2021flow, sanokowski2024diffusion, zhang2023let, bengio2021machine, karimi2022machine, mazyavkina2021reinforcement, vesselinova2020learning}, with different algorithms tailored to specific problem families. Reinforcement learning and graph neural networks are often used to train policies that construct solutions sequentially—for example, pointer networks for the Traveling Salesman Problem (TSP) \cite{bello2016neural}, and graph-embedding-based greedy heuristics for problems such as Minimum Vertex Cover, Max-Cut, and TSP \cite{khalil2017learning}.

Generative Flow Networks (GFlowNets) \cite{bengio2021flow} learn a generative policy whose stationary distribution is proportional to a user-defined reward function, allowing the model to sample diverse, high-quality solutions. GFlowNets have been applied across a range of combinatorial domains \cite{zhang2023let, jain2023gflownets}, and extensions have incorporated local search steps to improve the peak quality of generated solutions \cite{kim2023local}. More recently, works on temperature-conditioned GFlowNets\cite{zhang2023robust, kim2024learning, zhou2024phylogfn} has shown that explicitly learning or scaling temperature can stabilize training, enhance mode discovery, and improve generalization across diverse tasks.

\subsection{Neural Generators of Boltzmann Distributions}
In a parallel line of work, machine learning has been used to parameterize Boltzmann distributions, with the goal of accelerating sampling in MCMC-based algorithms. Early efforts focused on applications in statistical physics, where neural networks were trained to model physical systems \cite{wu2019solving} and improve thermodynamic simulations \cite{noe2019boltzmann, mcnaughton2020boosting}. More recently, Variational Neural Annealing (VNA) \cite{hibat2021variational} proposed using simulated annealing directly on the learned distribution for optimization. Diffusion-based models have also been explored in this context, where random noise is iteratively denoised into high-quality solutions \cite{sanokowski2024diffusion}, following a similar energy-guided generative philosophy.

\subsection{Summary}
Our work bridges ideas from generative modeling, variational annealing, and learning-based combinatorial optimization. Like Bayesian optimization, our method is sample-efficient and applicable to black-box objectives, but avoids the overhead of acquisition optimization. Compared to reinforcement learning and GFlowNet-based methods, our approach does not require problem-specific design choices, and instead learns a global, temperature-conditioned distribution over solutions. Finally, while prior work on neural Boltzmann samplers has focused either on physical simulations or annealing single distributions, our model learns a smooth family of energy-guided distributions across temperatures, unifying exploration and exploitation in a single, end-to-end framework.

\section{Methods}
We consider the minimization of a black-box Boolean function $f: \{0, 1\}^{n} \to \mathbb{R}$, where the input is a binary string $x$ of length $n$, and the output is a real-valued objective that we aim to minimize. The goal is to find the global optimum,
\begin{equation}
    x^* = \arg\min_x f(x).
\end{equation}
A naive approach would be to exhaustively evaluate all $2^n$ possible inputs, but this becomes computationally infeasible as $n$ increases.

Simulated annealing tackles this problem by sampling from the Boltzmann distribution:
\begin{equation}
    p(x, \beta) = \frac{e^{-\beta f(x)}}{Z(\beta)}, \label{eq:P}
\end{equation}
where $\beta$ is the inverse temperature, and $Z(\beta) = \sum_x e^{-\beta f(x)}$ is the partition function. Sampling is typically performed using Markov Chain Monte Carlo (MCMC). To locate the global minimum, the algorithm gradually increases $\beta$ from $0$ to a large value, causing $p(x, \beta)$ to evolve from a uniform distribution to a delta distribution peaked at $x^*$.

In theory, if the annealing schedule is sufficiently slow, the sampled distribution $\tilde{p}(x)$ remains close to the true $p(x)$, and the sampler converges to the global optimum. However, for highly correlated or frustrated energy landscapes, MCMC mixing can be exponentially slow, and convergence guarantees require annealing schedules that are themselves exponentially slow. In practice, simulated annealing often becomes trapped in local minima and fails to reach the global optimum efficiently.

GNA models the temperature-conditioned distribution $q_\theta(x|\beta)=\prod_{t=1}^n q_\theta(x_t|x_{<t},\beta)$ with an autoregressive, decoder-only transformer to approximate the Boltzmann distribution in Eq.~\eqref{eq:P}. Conditioning on the inverse temperature is implemented by projecting $\log \beta$ into the embedding space and adding this vector to all token embeddings.  At each annealing step, $\beta$ is set by the schedule and we draw i.i.d. candidates $x^{(k)}\sim q_\theta(\cdot|\beta)$; all samples are evaluated and retained as training data. This concentrates queries in high-probability regions while preserving stochastic exploration. Further numerical details are described in Appendix~\ref{appendix:methods}.

In the absence of prior knowledge about the problem structure, the transformer’s all-to-all attention mechanism provides a flexible and general architecture capable of modeling interactions between any pair of input variables. Stacking multiple transformer layers further enables the model to capture higher-order dependencies.

We consider two different problem settings:
\begin{enumerate}
    \item \textbf{Unlimited queries:} Each query to the black-box function $f$ is inexpensive relative to a single training step, and does not pose a computational bottleneck. This is the typical setting for many combinatorial optimization algorithms, where the challenge lies primarily in navigating the hardness of the solution space.
    \item \textbf{Limited queries:} Each query to $f$ is expensive compared to a training step, and the total number of queries is severely constrained. In this case, the goal is to find a high-quality solution with as few function evaluations as possible. This setting is common in scenarios like hyperparameter optimization, where each experiment (or query) may involve significant computational or real-world cost. Bayesian optimization is often used in such cases.
\end{enumerate}

While GNA achieves strong performance in both regimes, the training strategies must be adapted to the available query budget.

\subsection{Unlimited Queries}
In the unlimited-query setting, we adopt a standard approach used in prior works \cite{wu2019solving, hibat2021variational}, and train the model to approximate the Boltzmann distribution in Eq.~\eqref{eq:P} by minimizing the thermodynamic free energy, defined as $F = E - TS$. Here, the energy $E$ corresponds to the expected function value $f(x)$, $T = 1/\beta$ is the temperature, and $S$ is the entropy of the learned distribution.

Let $q_\theta(x, \beta)$ denote the neural network’s approximation to the Boltzmann distribution, where $\theta$ are the model parameters. For a fixed inverse temperature $\beta$, the free energy is given by:
\begin{equation}
    F(\beta) = \langle f(x) \rangle_{q_\theta} - \frac{1}{\beta} S(\beta),
\end{equation}
where the entropy is defined as
\begin{equation}
    S(\beta) = -\sum_x q_\theta(x, \beta) \log q_\theta(x, \beta), \label{eq:S}
\end{equation}
and $\langle \cdot \rangle_{p}$ denotes expectation with respect to distribution $p$.

The gradient of the free energy with respect to the model parameters can be estimated efficiently by sampling from $q_\theta$:
\begin{equation}
    \nabla_\theta F(\beta) = \Big\langle \big(F_{\text{loc}}(x, \beta) - \langle F_{\text{loc}}(x, \beta) \rangle_{q_\theta} \big) \nabla_\theta \log q_\theta(x, \beta) \Big\rangle_{q_\theta}, \label{eq:F_grad}
\end{equation}
where $F_{\text{loc}}(x, \beta) = f(x) + \frac{1}{\beta} \log q_\theta(x, \beta)$ is the local estimator of the free energy. This gradient estimator is known as the REINFORCE algorithm \cite{williams1992simple} in the context of reinforcement learning. See Appendix~\ref{appendix:F_grad} for a full derivation of Eq.~\eqref{eq:F_grad}.

\subsection{Limited Queries}
When evaluating $f(x)$ is expensive, sample efficiency becomes the primary concern, and each query must be used as effectively as possible. In this setting, we maintain a replay buffer containing all past evaluations $\{(x_i, f(x_i))\}$, and train the neural network to match the partial distributions:
\[
    \tilde{p}(x_i, \beta) = \frac{p(x_i, \beta)}{\sum_j p(x_j, \beta)} \quad \text{and} \quad 
    \tilde{q}_\theta(x_i, \beta) = \frac{q_\theta(x_i, \beta)}{\sum_j q_\theta(x_j, \beta)},
\]
defined over the observed samples $\{x_i\}$. The network is trained by minimizing the KL divergence between the true and approximate partial distributions, $D_{\text{KL}}(\tilde{p} \,\|\, \tilde{q}_\theta)$. This computation is inexpensive and straightforward, since it only involves samples already stored in the replay buffer.

We employ an active learning strategy to guide the query process. Training begins with 20 randomly selected queries to initialize the replay buffer. New query candidates are then selected by drawing small batches of samples from the model. To balance exploration and exploitation, the sampling temperature is gradually decreased following a predefined annealing schedule, akin to simulated annealing.

In this limited-query regime, overfitting is a major concern—since the replay buffer may be small, it is critical that the model generalizes beyond the observed samples rather than memorizing them. To mitigate this, we reserve $10\%$ of the samples as a validation set and monitor the validation loss during training. After every 20 new black-box queries, we revert the model to the version that achieved the lowest validation loss over the most recent 20-query window.

\section{Experiments}

We evaluate GNA on five combinatorial optimization problems: two relatively easy tasks (Ising sparsification \cite{baptista2018bayesian}, contamination control \cite{hu2010contamination}) and three hard problems (3-SAT \cite{barthel2002hiding}, 3-regular 3-XORSAT \cite{kowalsky20223}, and subset sum \cite{horowitz1974computing}). Detailed descriptions of each problem are provided in Appendix~\ref{appendix:problems}. Below, we briefly outline the black-box objective $f(x)$ for each case:

\begin{itemize}
    \item \textbf{Ising sparsification:} Remove interactions from an Ising spin-glass while minimally perturbing its statistical behavior. The objective $f(x)$ is the KL divergence between the original and sparsified spin distributions, with an added regularization term.
    
    \item \textbf{Contamination control:} Optimize prevention strategies in a food supply chain to minimize overall cost and contamination risk. Here, $f(x)$ combines the cost of preventive measures with a penalty for violating contamination thresholds.
    
    \item \textbf{3-SAT:} Barthel instances \cite{barthel2002hiding} with planted solutions at a clause-to-variable ratio of 4.3, near the known complexity peak. The objective $f(x)$ is the number of unsatisfied clauses.
    
    \item \textbf{3-XORSAT:} Randomly generated 3-regular 3-XORSAT instances \cite{kowalsky20223}, where each variable appears in exactly three clauses. The objective $f(x)$ is again the number of unsatisfied clauses.
    
    \item \textbf{Subset sum:} Hard instances generated at the critical density where $n = L$, the number of binary digits needed to represent the target sum \cite{horowitz1974computing}. The objective is defined as
    \[
        f(x) = \log \left(\text{diff. between computed and target sum} + 1 \right)    \]
\end{itemize}

All problems are evaluated in a black-box setting, where the solver has access only to function evaluations of the target objective. Each problem is tested under the limited-query scenario, while the three harder problems (3-SAT, 3-XORSAT, and subset sum) are additionally evaluated under the unlimited-query setting.

We benchmark GNA against several established black-box optimizers: Simulated Annealing (SA) \cite{kirkpatrick1983optimization}, Tree-structured Parzen Estimator (TPE) \cite{bergstra2013making}, Bayesian Optimization of Combinatorial Structures (BOCS) \cite{baptista2018bayesian}, and COMBO \cite{oh2019combinatorial}.

We evaluate two variants of GNA, based on different annealing strategies. In the first variant, which we refer to as \textbf{GNA-SA}, the temperature is decreased monotonically throughout training, following a schedule similar to that of simulated annealing. In the second variant, denoted \textbf{GNA-PT}, a temperature is randomly sampled at each training step from a predefined interval. The minimum of this interval gradually decreases over time. This mimics the behavior of parallel tempering (PT), which maintains multiple replicas of the system at different temperatures and facilitates transitions between modes.

\subsection{Limited queries}

\begin{table*}
  \caption{Final objective values achieved on each benchmark problem under a 200-query budget (mean $\pm$ standard deviation over 10 runs).}
  \label{tab:limited}
  \centering
  \begin{tabular}{cccccc}
    \toprule
    Method   & Ising & Contamination & 3-SAT & 3-XORSAT & Subset sum \\
    \midrule
    SA       & $0.61 \pm 0.51$ & $21.72 \pm 0.14$ & $1.80 \pm 0.75$ & $3.40 \pm 0.49$ & $12.32 \pm 0.98$ \\
    TPE      & $0.82 \pm 0.60$ & $21.62 \pm 0.10$ & $3.80 \pm 1.25$ & $5.40 \pm 1.28$ & $12.74 \pm 1.06$ \\
    BOCS-SDP & $0.38 \pm 0.10$ & $21.65 \pm 0.06$ & $2.50 \pm 0.50$ & $5.90 \pm 0.70$ & $12.17 \pm 1.19$ \\
    BOCS-SA  & $0.34 \pm 0.11$ & $21.63 \pm 0.15$ & $1.60 \pm 0.49$ & $6.20 \pm 1.17$ & $12.39 \pm 0.74$ \\
    COMBO    & $\mathbf{0.21 \pm 0.01}$ & $\mathbf{21.38 \pm 0.14}$ & N/A\hyperlink{tablefootnote}{$^a$} & N/A\hyperlink{tablefootnote}{$^a$} & N/A\hyperlink{tablefootnote}{$^a$} \\
    \midrule
    GNA-SA   & $0.24 \pm 0.04$ & $21.39 \pm 0.10$ & $1.60 \pm 1.02$ & $3.30 \pm 1.41$ & $11.86 \pm 2.00$ \\
    GNA-PT   & $0.30 \pm 0.16$ & $21.41 \pm 0.14$ & $\mathbf{1.20 \pm 0.87}$ & $\mathbf{3.20 \pm 1.25}$ & $\mathbf{11.38 \pm 0.99}$ \\
    \bottomrule
    \multicolumn{6}{l}{\footnotesize{\hypertarget{tablefootnote}{$^a$} Not evaluated due to lack of support for user-defined black-box functions in COMBO.}}
  \end{tabular}
\end{table*}

\begin{figure}
    \centering
    \includegraphics[width=0.98\linewidth]{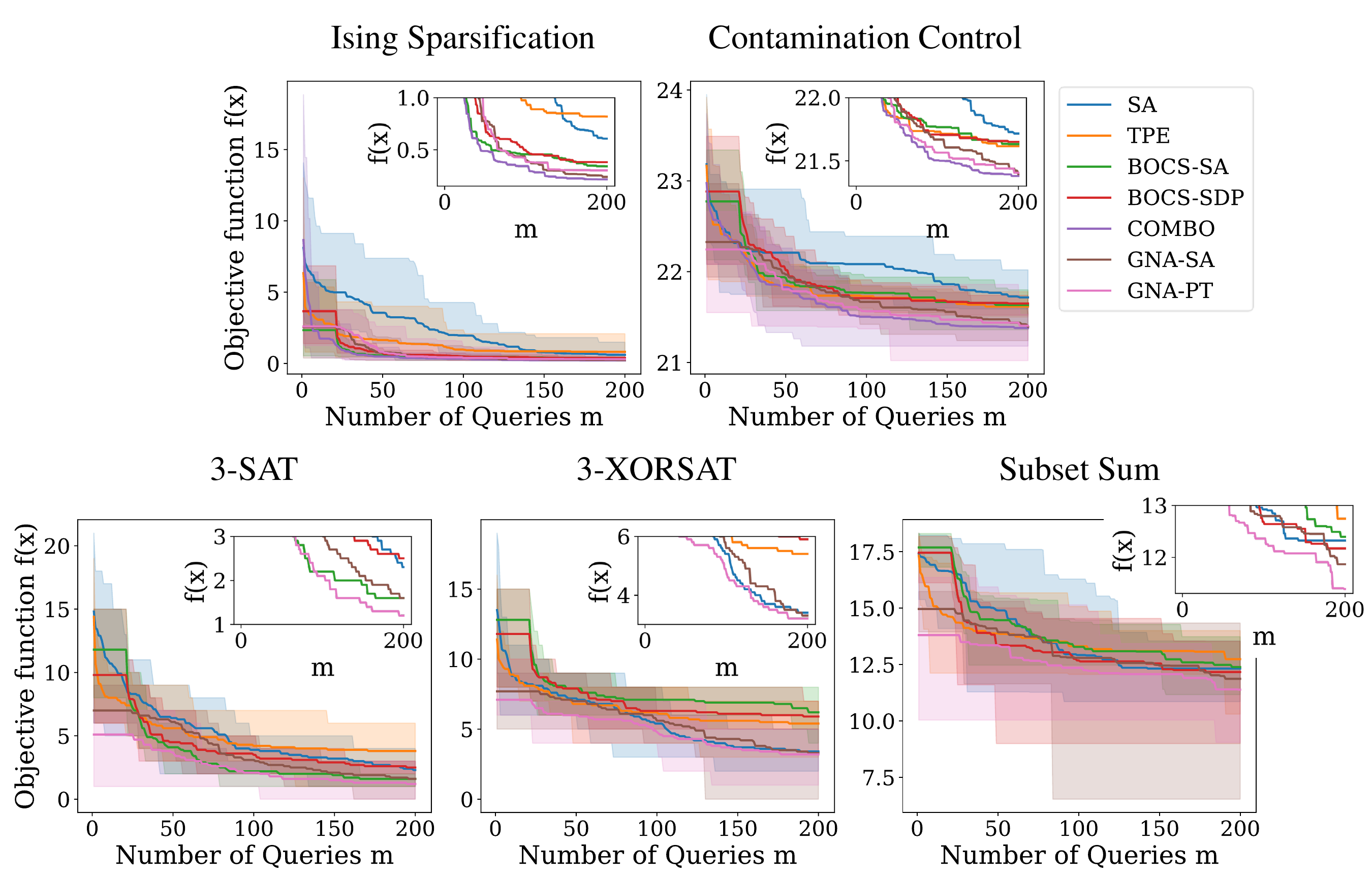}
    \caption{Best objective value found so far as a function of the number of black-box queries $m$, across all benchmark problems and solvers. Solid curves denote the mean over 10 independent runs, and shaded regions span the full range (minimum to maximum) of values observed. Insets show zoomed-in views near the best-found solutions.}
    \label{fig:limited_query}
\end{figure}

In this setting, we fix the number of variables to $n = 25$ for contamination control, 3-SAT, 3-XORSAT, and subset sum, and $n = 24$ for Ising sparsification. Each solver is allowed a maximum of 200 queries to the black-box function, with 20 initial random samples used to initialize BOCS, COMBO, and GNA. All experiments are repeated across 10 independent runs. Table~\ref{tab:limited} reports the mean $\pm$ standard deviation of the best objective value found, while Fig.~\ref{fig:limited_query} plots the minimum objective as a function of query count. Solid curves show the mean performance, and shaded regions represent the range across runs.

For the two easier problems, the energy landscape is relatively smooth and low-energy solutions tend to be close to one another. COMBO achieves the best overall performance, with GNA-SA closely matching it. However, each run of COMBO requires approximately 8 hours of computation on 48 CPU cores\footnote{AMD EPYC 7401}, while GNA-SA completes it in about 20 seconds on a single GPU\footnote{NVIDIA TITAN RTX}. Interestingly, GNA-PT underperforms in this scenario, likely because both problems have narrow basins of attraction around their optima, making low-temperature ``exploitation'' more effective than high-temperature ``exploration.''

For the three harder problems, the objective landscapes are rugged, with many suboptimal local minima and only a single global optimum. In this regime, GNA-PT performs best overall. We also observe that vanilla SA performs competitively on 3-SAT and 3-XORSAT, likely because a sequence of single-bit flips can quickly descend to a local minimum that is close in energy to the global optimum. Notably, GNA is the only method that successfully identifies the global optimum at least once in both 3-SAT and 3-XORSAT. In the subset sum problem, the optimal value is $f(x) = 0$, but none of the algorithms—including GNA—are able to reach a solution close to this value, underscoring the intrinsic hardness of this benchmark.

\subsection{Unlimited queries}

We now evaluate GNA-PT in the unlimited-query setting on the three harder problems: 3-SAT, 3-XORSAT, and subset sum. In this regime, the neural network is trained using the free energy gradient (Eq.~\eqref{eq:F_grad}), and queries to the black-box function are no longer restricted. For each problem, we gradually increase the number of variables $n$ and record the number of training steps required to find the optimal solution (i.e., when $f(x) = 0$).

\begin{figure}
    \centering
    \includegraphics[width=0.98\linewidth]{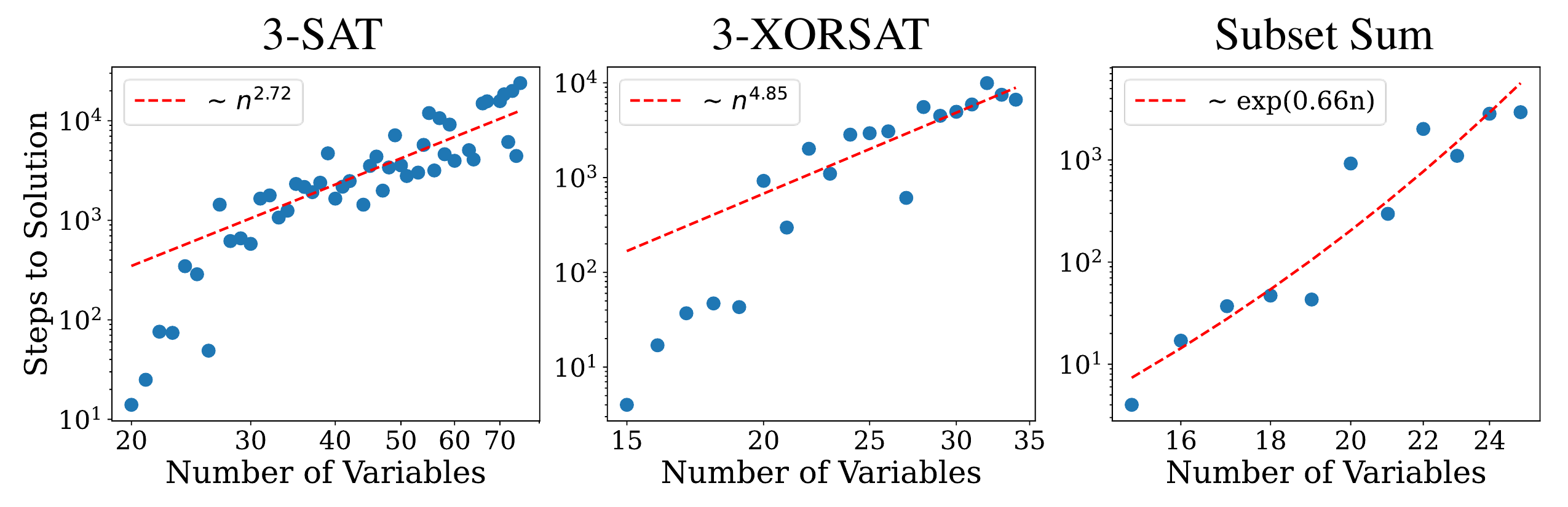}
    \caption{Number of training steps required to reach the optimal solution in the unlimited-query setting. Data points show the median over three independent runs. The red dashed curve indicates a best-fit scaling for the solvable regime.}
    \label{fig:unlimited_query}
\end{figure}

To limit runtime, we set an upper bound on the number of training steps: $n^3/8$ for 3-SAT, and $10^4$ for both 3-XORSAT and subset sum. If the model does not reach the optimal solution within this limit, the run is considered failed. We repeat each experiment three times and report the median number of steps in Fig.~\ref{fig:unlimited_query}.

Among the three problems, 3-SAT is the most tractable in this setting: GNA-PT is able to solve instances up to $n = 75$, with an empirical scaling of $n^{2.72}$. While 3-XORSAT is solvable in polynomial time when the problem structure is known \cite{kowalsky20223}, it proves far more challenging in the black-box setting, with GNA exceeding the step limit for problem sizes as small as $n \sim 35$. Subset sum remains the hardest: even with unlimited queries, GNA does not significantly outperform random search, despite achieving the best performance among all methods in the limited-query setting.

\section{Analysis: Opening the black box}

\begin{figure}
    \centering
    \includegraphics[width=0.98\linewidth]{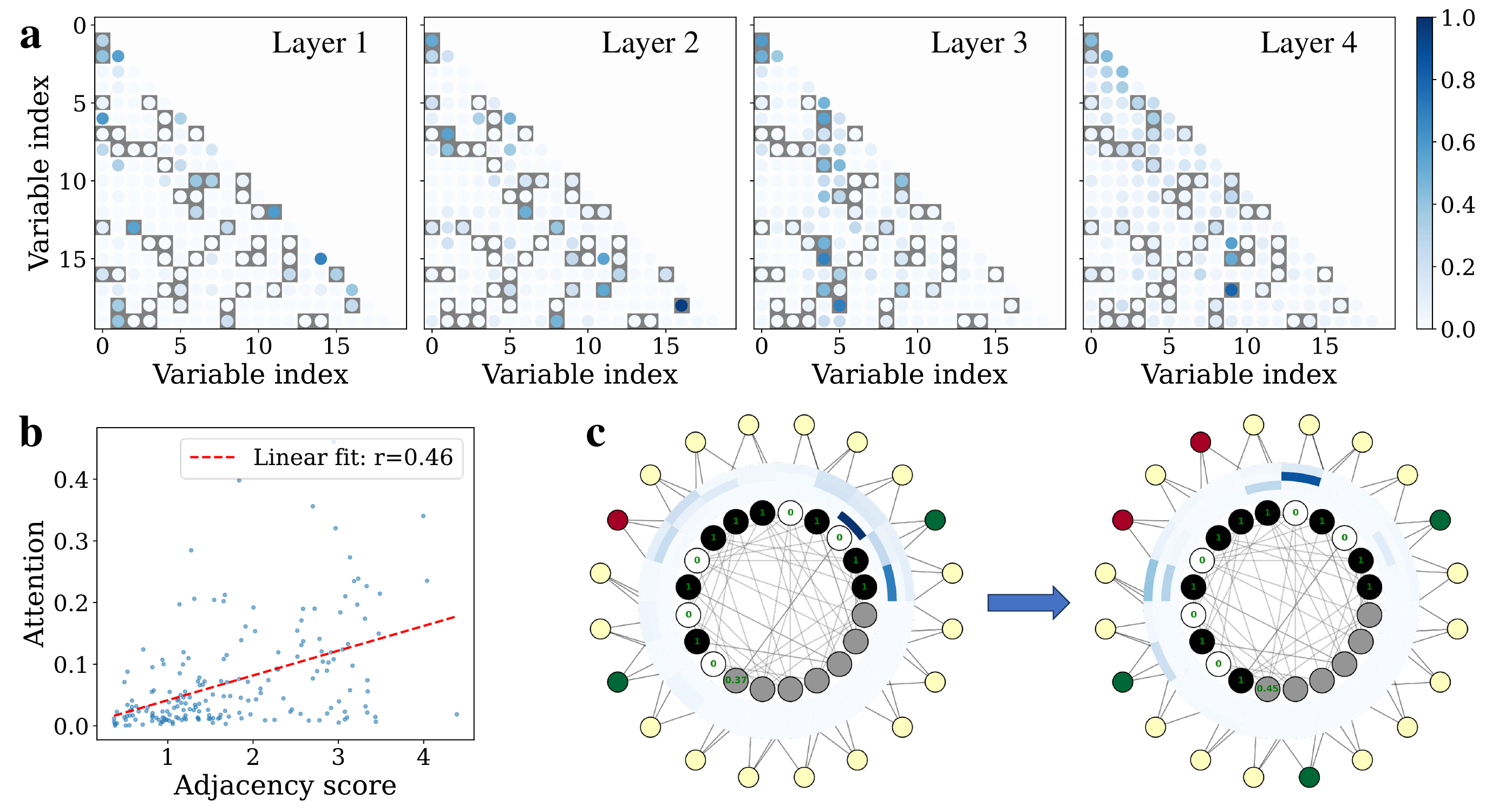}
    \caption{Visualization of the attention mechanism learned by GNA on a 3-XORSAT instance with 20 variables. All results are obtained at inverse temperature $\beta = 1$, averaged over 1000 samples. (a) Attention maps from each of the four transformer layers. The background grid encodes variable co-occurrence in clauses (darker indicates a shared clause), and the circles represent pairwise attention values. (b) Scatter plot of attention values versus adjacency scores defined in Eq.~\eqref{eq:adj}, showing a positive correlation. (c) A single inference step by GNA. Inner circles represent variable states (white: 0, black: 1, gray: undetermined), and outer circles represent clause satisfaction (green: satisfied, red: unsatisfied, yellow: undetermined). Variables are connected if they share a clause; the multi-ring structure in the center encodes attention weights for the variable to be predicted. After sampling, the updated configuration and clause states are shown.}
    \label{fig:attention}
\end{figure}

GNA is able to solve hard combinatorial optimization problems without access to explicit problem structure. This raises a natural question: \textit{what does the model actually learn from black-box queries}? To investigate this question, we visualize the attention maps learned by GNA when trained on a 20-variable 3-XORSAT instance under the unlimited-query setting, and compare them to the underlying structure of the problem.

The model architecture consists of 4 transformer layers, each with hidden size 32 and single-head attention. We average the attention maps over a batch of 1000 samples drawn at temperature $\beta = 1$. The resulting attention matrices from each of the 4 layers are shown in Fig.~\ref{fig:attention}(a). Remarkably, GNA appears to capture the geometry of the problem: variables that are structurally closer in the constraint graph attend more strongly to each other.

To quantify this relationship, we define an \textit{adjacency score} matrix $S$ based on the problem's graph structure:
\begin{equation}
    S = \sum_{k=1}^{l} \alpha^{k-1} A^k, \label{eq:adj}
\end{equation}
where $A$ is the adjacency matrix of the problem, $\alpha$ is a discount factor satisfying $0 < \alpha < 1/|\lambda_{\max}|$ (with $\lambda_{\max}$ the largest eigenvalue of $A$), and $l$ is the maximum path length considered. The $(i,j)$-th entry of $S$ counts all paths of length up to $l$ between variables $i$ and $j$, with shorter paths weighted more heavily. In our analysis, we use $\alpha = 0.15$ and $l = 10$.

In Fig.~\ref{fig:attention}(b), we plot the average attention between variable pairs against their corresponding adjacency scores from Eq.~\eqref{eq:adj}. A linear fit yields a Pearson correlation coefficient of $r = 0.46$, indicating a significant relationship between the learned attention patterns and the underlying problem connectivity.

Finally, in Fig.~\ref{fig:attention}(c), we illustrate both the problem structure and a single step of inference performed by GNA. The inner circles represent Boolean variables (white for 0, black for 1, and gray for undetermined), while the outer circles correspond to clauses (green for satisfied, red for unsatisfied, and yellow for undetermined). Edges between variables indicate their co-occurrence in a clause. The number displayed on the first undetermined variable indicates the model's predicted probability that the next bit is 1—in this case, for the left panel, 37\%.

The blue concentric rings depict the attention weights from the current step (colormap shared with Fig.~\ref{fig:attention}(a)). Notably, the model assigns higher attention to variables that are directly connected to the one being predicted, reflecting awareness of the underlying constraint structure. The right panel shows the configuration after sampling this variable; although 1 was the less likely outcome, it was selected during sampling, leading to one additional unsatisfied clause. This behavior is expected at $\beta = 1$, a relatively high temperature in the exploration phase, where suboptimal configurations are still frequently sampled.

\section{Conclusions}

We introduced GNA, a flexible and general framework for solving black-box combinatorial optimization problems by performing annealing over a learned, temperature-conditioned distribution. GNA achieves strong performance across a diverse set of benchmarks, competing effectively with state-of-the-art black-box optimizers.

Despite its strengths, GNA also has several limitations. First, in the limited-query setting, its performance can be sensitive to the initial samples, leading to variability across runs—as reflected in the relatively large standard deviations reported in Table~\ref{tab:limited}. Second, GNA's effectiveness depends on careful design of the annealing schedule and other hyperparameters. While we use fixed schedules across problems in our experiments, optimal performance in practice may require domain-specific tuning. Finally, overfitting remains a concern when training on small datasets. Our validation-based early stopping strategy helps mitigate this issue, but more advanced regularization techniques could further improve robustness.

While GNA excels in the black-box setting, it is inherently modular and extensible. In practice, many real-world optimization problems come with additional structure—such as known constraints, domain knowledge, or access to existing solvers. GNA can naturally incorporate this information, and we envision a broader paradigm: using generative models to learn the flow fields or internal dynamics of established optimization algorithms, such as memcomputing solvers \cite{di2022memcomputing} for NP problems, conflict-driven clause learning (CDCL) \cite{silva1996grasp} for SAT, or hardware-inspired approaches like quantum annealing \cite{kadowaki1998quantum} and Ising machines \cite{mohseni2022ising}.

By modeling the distribution of optimization trajectories generated by such solvers, GNA could serve as a foundation for a meta-optimization framework—one that not only solves combinatorial problems effectively, but also adapts and improves over time by learning from the behavior of existing methods.

\section*{Acknowledgments}
Y.-H. Zhang and M. Di Ventra are supported by NSF grant No. ECCS-2229880. A full implementation of the GNA algorithm can be found at \url{https://github.com/yuanhangzhang98/generative_neural_annealer}.

{\appendices

\section{Numerical details}
\label{appendix:methods}
This section provides the implementation and training details necessary to reproduce our results.

Our model adopts a standard decoder-only, autoregressive transformer architecture \cite{vaswani2017attention}, with two distinct input tokens representing 0 and 1. Because the position of each variable does not necessarily reflect any geometric structure, the positional embeddings are fully learned. To condition the model on inverse temperature $\beta$, we introduce an additional temperature embedding: a simple linear layer that projects $\log \beta$ into the same embedding space. During inference, a 0-token is prepended to the input sequence to initiate autoregressive generation.

In the limited-query setting, the model consists of 3 transformer layers with single-head attention and a hidden size of 20. In the unlimited-query setting, the model has 4 layers, each with single-head attention and a hidden size of 32. Training is performed on a single NVIDIA TITAN RTX GPU with 24GB VRAM.

To achieve optimal performance, the annealing schedule is treated as a set of hyperparameters. In the limited-query regime, we set $\beta_{\min} = 0.057$ and $\beta_{\text{upper\_bound}} = 69.7$. Initially, $\beta_{\max}$ is set equal to $\beta_{\min}$ and then gradually increases to $\beta_{\text{upper\_bound}}$ over the first 33\% of queries, increasing linearly in log-space. For GNA-SA, both training and sampling are always performed at $\beta_{\max}$. For GNA-PT, each training step samples $\beta$ uniformly from $[\beta_{\min}, \beta_{\max}]$, while queries are generated at $\beta_{\max}$.

Training is performed using the AdamW optimizer \cite{loshchilov2017decoupled} with a learning rate of $8.2 \times 10^{-4}$ and weight decay of $1.5 \times 10^{-4}$. These hyperparameters are tuned using the HyperOpt library \cite{bergstra2013making} and kept fixed across all limited-query experiments.

Each run is initialized with 20 random queries to the black-box function, with 18 used for training and 2 reserved for validation. After each new query, the model is trained for 5 steps in GNA-SA and 25 steps in GNA-PT. New samples are added to the training set with probability 0.9 and to the validation set otherwise. However, if the new sample improves upon the current best solution, it is always added to the training set. We monitor validation loss throughout training and, after every 20 queries, revert the model to the checkpoint with the lowest validation loss from the most recent 20-query window.

In the unlimited-query setting, we set $\beta_{\min} = 0.1$ and $\beta_{\text{upper\_bound}} = 100$. Training begins with $\beta_{\max} = 1$, which is gradually increased to $\beta_{\text{upper\_bound}}$ over $2 \times 10^4$ training steps, following a linear schedule in log space. Training is performed using the free-energy gradient (Eq.~\eqref{eq:F_grad}) paired with the Adam optimizer \cite{kingma2014adam} with a fixed learning rate of $5 \times 10^{-4}$. These hyperparameters are held constant across all experiments in the unlimited-query regime and have not been fine-tuned; we expect further performance improvements with additional tuning.

The expectation in Eq.~\eqref{eq:F_grad} is estimated using $N_{\text{unique}} = 10^3$ unique configurations, along with a sample reweighting strategy proposed in \cite{barrett2022autoregressive}. Specifically, we first generate partial sequences of variables up to length $k$, denoted $\mathbf{x}^k = x_1x_2\cdots x_k$, using a large batch size of $N_{\text{batch}} = 10^6$. We record the number of occurrences of each unique $\mathbf{x}^k$ until the number of unique partial sequences exceeds $N_{\text{unique}}$. At that point, we stop generating new branches and complete each partial sequence by sampling the remaining variables autoregressively using the standard method.

This two-stage strategy allows us to simulate a much larger effective batch size while only evaluating a modest number of unique configurations. As a result, we improve computational efficiency and reduce the variance in the estimated gradients.

\section{Free energy gradient}
\label{appendix:F_grad}
Here, we derive the expression for the free energy gradient used in Eq.~\eqref{eq:F_grad} of the main text.

We begin with the definition of the free energy:

\begin{equation}
    F(\beta) = \langle f(x)\rangle_{q_\theta(x, \beta)} + \frac{1}{\beta}\sum_x q_\theta(x, \beta)\log q_\theta(x, \beta)\label{eq:F_supp}
\end{equation}

To compute the gradient $\nabla_\theta F(\beta)$ with respect to the model parameters $\theta$, we rewrite Eq.~\eqref{eq:F_supp} as:

\begin{equation}
\begin{aligned}
    F(\beta) =& \sum_x q_\theta(x, \beta)\big(f(x) + \log q_\theta(x, \beta)/\beta\big)\\
        =& \langle F_\text{loc}(x, \beta)\rangle_{q_\theta} \label{eq:Floc}
\end{aligned}
\end{equation}
where we define the local free energy estimator as
\begin{equation}
    F_{\text{loc}}(x, \beta) = f(x) + \frac{1}{\beta} \log q_\theta(x, \beta).
\end{equation}

We then take the gradient with respect to $\theta$:
\begin{equation}
\begin{aligned}
    \nabla_\theta F(\beta) =& \sum_x\Big(F_\text{loc}(x, \beta)\nabla_\theta q_\theta(x, \beta)+ q_\theta(x, \beta)\nabla_\theta F_\text{loc}(x, \beta)\Big)    \\
    =& \langle F_\text{loc}(x, \beta)\nabla_\theta\log q_\theta(x, \beta)\rangle_{q_\theta} + \langle\nabla_\theta F_\text{loc}(x, \beta)\rangle_{q_\theta} \label{eq:F_grad_expanded}
\end{aligned}
\end{equation}

Using the identity
\begin{equation}
\begin{aligned}
    \langle\nabla_\theta \log q_\theta (x, \beta)\rangle_{q_\theta}=&\sum_x q_\theta(x, \beta)\cdot \frac{1}{q_\theta(x, \beta)}\nabla_\theta q_\theta(x, \beta)\\
    =&\nabla_\theta\sum_x q_\theta (x, \beta)=\nabla_\theta 1=0,
\end{aligned}
\end{equation}
the second term in Eq.~\eqref{eq:F_grad_expanded} vanishes. Moreover, we can subtract the following zero-mean baseline to reduce the variance of the gradient estimator \cite{goodfellow2016deep}:
\begin{equation}
    0 = \langle F_{\text{loc}}(x, \beta) \rangle_{q_\theta} \cdot \langle \nabla_\theta \log q_\theta(x, \beta) \rangle_{q_\theta}.
\end{equation}
This gives the final expression for the free energy gradient:
\begin{equation}
    \nabla_\theta F(\beta) = \Big\langle \Big(F_\text{loc}(x, \beta)-\langle F_\text{loc}(x, \beta)\rangle_{q_\theta}\Big)\nabla_\theta\log q_\theta(x, \beta)\Big\rangle_{q_\theta} \label{eq:F_grad_supp}
\end{equation}

This expression can be efficiently estimated by drawing samples from the model distribution $q_\theta(x, \beta)$. Importantly, the local free energy estimator $F_{\text{loc}}(x, \beta)$ satisfies a zero-variance property when $q_\theta$ exactly matches the target Boltzmann distribution $p(x, \beta) = e^{-\beta f(x)} / Z(\beta)$:
\begin{equation}
\begin{aligned}
    F_{\text{loc}}(x, \beta)
    &= f(x) + \frac{1}{\beta} \log p(x, \beta) \\
    &= f(x) + \frac{1}{\beta} (-\beta f(x) - \log Z(\beta)) \\
    &= -\frac{1}{\beta} \log Z(\beta),
\end{aligned}
\end{equation}
which is independent of $x$. Therefore, as $q_\theta$ approaches the target distribution, the variance of the estimator decreases, and gradient estimates become more accurate.

\section{Benchmark problems}
\label{appendix:problems}

Here, we provide detailed descriptions of all benchmark problems used in the main text.

\subsection{Ising Sparsification}  
This toy problem, first introduced in \cite{baptista2018bayesian}, has been used as a benchmark for various Bayesian optimization methods \cite{baptista2018bayesian, oh2019combinatorial, dadkhahi2020combinatorial}. An Ising spin glass is defined on a square lattice with $z \in \{-1, 1\}^n$ and distribution
\[
    p(z) = \frac{1}{Z_p} \exp\left( \sum_{ij} J_{ij} z_i z_j \right),
\]
where each coupling $J_{ij}$ is drawn uniformly from $[0.05, 5]$, and its sign is chosen at random with probability $1/2$.

The goal is to sparsify the model by removing as many couplings $J_{ij}$ as possible, while preserving the original distribution $p(z)$ as closely as possible. Let $x_{ij} \in \{0, 1\}$ denote a binary decision variable indicating whether the edge $J_{ij}$ is retained. The sparsified model is defined as
\[
    q_x(z) = \frac{1}{Z_q} \exp\left( \sum_{ij} x_{ij} J_{ij} z_i z_j \right),
\]
and the objective function is
\begin{equation}
    f(x) = D_\mathrm{KL}(p \,\|\, q_x) + \lambda \|x\|_1,
\end{equation}
where $\lambda$ is a regularization parameter that controls the trade-off between sparsity and fidelity to the original model. In our experiments, we set $\lambda = 0.01$.

\subsection{Contamination Control}  
The contamination control problem \cite{hu2010contamination} has been widely used as a benchmark for Bayesian optimization methods \cite{baptista2018bayesian, oh2019combinatorial, dadkhahi2020combinatorial}. In our experiments, we follow the exact setup described in \cite{baptista2018bayesian}.

The scenario models a food supply chain with $n$ processing stages. At each stage $i$, the fraction of contaminated product, denoted $Z_i$, evolves based on whether or not a preventive action is taken. The binary decision variable $x_i \in \{0, 1\}$ indicates whether prevention is performed at stage $i$ ($x_i = 1$ for prevention). The goal is to minimize the total cost of prevention while keeping contamination levels under control. The objective function is defined as:
\begin{equation}
    f(x)=\sum_{i=1}^n \left( c_i x_i + \rho \big( \Theta(Z_i - U_i) - \epsilon \big) \right), \label{eq:contamination_goal}
\end{equation}
where $c_i = 1$ is the cost of prevention, $U_i = 0.1$ is the contamination threshold, $\rho = 1$ is the penalty for violations, $\epsilon = 0.05$ is a tolerance factor, and $\Theta(\cdot)$ is the unit step function.

The contamination dynamics follow:
\begin{equation}
    Z_i = \Lambda_i (1 - x_i)(1 - Z_{i-1}) + (1 - \Gamma_i x_i) Z_{i-1}, \label{eq:contamination_Z}
\end{equation}
where $Z_0 \sim \text{Beta}(1, 30)$ is the initial contamination level, and $\Lambda_i \sim \text{Beta}(1, 17/3)$, $\Gamma_i \sim \text{Beta}(1, 3/7)$ are random variables sampled independently for each stage. The full dynamics in Eq.~\eqref{eq:contamination_Z} are simulated 100 times, and the objective in Eq.~\eqref{eq:contamination_goal} is averaged across simulations to compute the final cost.

Some prior works \cite{baptista2018bayesian, oh2019combinatorial} included an additional $\ell_1$ regularization term $\lambda \|x\|_1$ in the objective. However, since this is equivalent to increasing the prevention cost $c_i \rightarrow c_i + \lambda$, we omit this term in our experiments.

\subsection{Barthel Instances of 3-SAT}  
The Boolean satisfiability (SAT) problem asks whether there exists an assignment of Boolean variables that satisfies a given Boolean formula. In the 3-SAT variant, the formula is composed of clauses, each formed by a logical OR of three (possibly negated) variables.

It is well known that random 3-SAT instances exhibit a phase transition from satisfiable to unsatisfiable as the clause-to-variable ratio $\alpha$ increases. This transition occurs near $\alpha = 4.27$, and instances near this threshold are empirically the hardest to solve \cite{hartmann2004new}. When $\alpha$ is much smaller or larger than this critical value, more efficient algorithms can often find solutions or refutations quickly.

The Barthel instances \cite{barthel2002hiding} are specially constructed 3-SAT problems designed to remain hard despite being satisfiable. They are generated using a statistical mechanics-inspired procedure. First, a planted solution is chosen to guarantee satisfiability. Then, clauses are added randomly in a way that respects the planted solution while maintaining an average local field of zero. This ensures that local search algorithms do not gain exploitable information from the clause structure, making the instances particularly challenging.

In our experiments, we benchmark solvers on Barthel instances with a clause-to-variable ratio $\alpha = 4.3$, slightly above the phase transition, to ensure the generated problems are both satisfiable and computationally difficult. The objective function $f(x)$ is the number of unsatisfied clauses, and the global minimum corresponds to $f(x) = 0$ where all clauses are satisfied. 

\subsection{3-Regular 3-XORSAT}  
The $k$-XORSAT problem is a variant of satisfiability in which each clause is an exclusive OR (XOR) of $k$ Boolean variables. In the 3-XORSAT case, each clause enforces a linear parity constraint of the form $x_i \oplus x_j \oplus x_k = b$, where $b \in \{0,1\}$ is a fixed parity bit and $\oplus$ denotes addition modulo 2. Unlike 3-SAT, XORSAT is linear over $\mathbb{F}_2$ and can be solved in polynomial time using Gaussian elimination when the full constraint matrix is available.

However, in the black-box setting, where only function evaluations (e.g., number of unsatisfied clauses) are available and no explicit structure is exposed, XORSAT remains a challenging optimization problem—particularly when the clause structure is adversarially chosen.

In our experiments, we use 3-regular 3-XORSAT instances \cite{kowalsky20223}, where each variable appears in exactly three clauses and each clause involves exactly three variables. This regularity ensures uniform constraint participation and maximizes frustration in the system, making local search difficult. Each instance is generated by randomly generating a planted solution and randomly selecting clauses subject to the regularity condition, while assigning the parity bits according to the planted solution. The objective function $f(x)$ is also the number of unsatisfied clauses.

\subsection{Subset Sum}  
The subset sum problem is a classical NP-complete problem: given a set of $n$ positive integers $\{a_1, a_2, \dots, a_n\}$ and a target sum $T$, the goal is to determine whether there exists a binary vector $x \in \{0, 1\}^n$ such that the selected subset sums exactly to $T$, i.e.,
\[
    \sum_{i=1}^n a_i x_i = T.
\]
In the decision version, the task is to determine whether such a subset exists; in the optimization version used here, the goal is to minimize the discrepancy between the subset sum and the target.

We generate instances near the so-called hardness peak, which occurs when the number of bits needed to represent $T$ (denoted $L$) is equal to the number of elements $n$ in the set \cite{horowitz1974computing}. In this regime, exhaustive search becomes infeasible, and no known pseudo-polynomial-time algorithms are effective in the general case.

To construct problem instances, we sample $n$ integers uniformly at random in the range $[1, 2^L]$ with $L = n$. The target sum $T$ is generated by randomly selecting a binary vector $x^* \in \{0,1\}^n$ and computing $T = \sum_i a_i x^*_i$, ensuring the instance has at least one known solution.

The objective function used for optimization is defined as:
\[
    f(x) = \log\left( \left| \sum_{i=1}^n a_i x_i - T \right| + 1 \right),
\]
which penalizes deviation from the target and smooths the optimization landscape. The global minimum $f(x) = 0$ is achieved if and only if the subset sum exactly equals $T$.

}

\bibliographystyle{IEEEtran}
\bibliography{bibliography}

\vfill

\end{document}